\documentclass[a4paper, 11pt]{article}
\usepackage[T1]{fontenc}
\usepackage[latin1]{inputenc}
\usepackage{stmaryrd}
\usepackage{amssymb}
\usepackage{graphics}
\usepackage{verbatim}
\usepackage{hyperref}

\usepackage{xcolor}
\definecolor{blue}{rgb}{.255,.41,.884} % RoyalBlue of svgnames
\definecolor{red}{rgb}{1,0,0} % Red of svgnames
\definecolor{green}{rgb}{.136,.704,.136} % LimeGreen of svgnames
\definecolor{yellow}{rgb}{1,.648,0} % Orange of svgnames

\providecommand{\keywords}[1]
{
  \textbf{\textit{Keywords ---}} #1
}

\title{The Knowledge-Behaviour Disconnect in LLM-based Chatbots\\ (pre-print)}

\author{Jan Broersen}

\begin{document}

\maketitle

\begin{abstract}

Large language model-based artificial conversational agents (like ChatGPT) give answers to all kinds of questions, and often enough these answers are correct. Just on the basis of that capacity alone, we may attribute knowledge to them. But do these models use this knowledge as a basis for their own conversational behaviour? I argue this is not the case, and I will refer to this failure as a `disconnect'. I further argue this disconnect is fundamental in the sense that with more data and more training of the LLM on which a conversational chatbot is based, it will not disappear. The reason is, as I will claim, that the core technique used to train LLMs does not allow for the establishment of the connection we are after. The disconnect reflects a fundamental limitation on the capacities of LLMs, and explains the source of hallucinations. I will furthermore consider the ethical version of the disconnect (ethical conversational knowledge not being aligned with ethical conversational behaviour), since in this domain researchers have come up with several additional techniques to influence a chatbot's behaviour. I will discuss how these techniques do nothing to solve the disconnect and can make it worse. 

\end{abstract}


\newpage

\tableofcontents

\newpage

\section{Introduction}
\label{Introduction}
In November 2022, ChatGPT became the first large language model-based artificial conversational agent (LLM-based Chatbot, for short) that was released to the wider public. LLM-based chatbots give answers to all kinds of questions, and often enough these answers are correct. Just on the basis of that capacity alone, we may attribute \emph{knowledge} to them. But do these AI systems \emph{use} this knowledge as a basis and source for their own conversational behaviour, like we humans do? In other words, do these agents show actual knowledge-based agency in the sense traditionally defined in AI? Or is it the case that their behaviour is \emph{disconnected} from the knowledge that is at display in the answers they give, thereby maybe implying that we might be much too quick to attribute a certain level of knowledge-based intelligent agency to them? 

This article will investigate to what extent the conversational behaviour displayed by LLM-based chatbots is disconnected from the knowledge we attribute to them on the basis of their apparent verbal competency. I will argue that, currently, there is such a disconnect and that it is particularly salient when it comes to ethical knowledge and prudent conversational behaviour. I will further argue that the disconnect is fundamental in the sense that it is different from anecdotal, similarly disconnected behaviour of human agents, and that it cannot be eliminated with current AI techniques. 

To argue for my points, I will first need to explain, in non-technical terms, how LLM-based chatbots work.  

\section{How LLM-based chatbots work}
\label{How LLM-based chatbots work}
%\smallskip \noindent [{\bf The LLM at the heart of ChatGPT}]\\ 
At their heart, systems like ChatGPT have a so-called large language model\footnote{I will not go into the question if it is correct use of established mathematical and  general scientific terminology to call the trained artificial neural nets of large language models, `models' of language. I will only say that I am reluctant to refer to trained neural nets as models, because in my opinion a model \emph{presupposes} a language. } (LLM) that is pre-trained on enormous amounts of textual data. The target function being learned is one that gives possible continuations of token/word sequences, and estimated likelihood scores for such continuations on the basis of all the continuation data in the input texts. The text continuations it gives back as answers to the questions you pose it, are then a combinatorial selection from the continuations that are most likely on the basis of all the texts the system has `seen' in its training data.\footnote{Likelihoods of continuations are not calculated in a statistically rigorous way, but with neural nets. In the early days of language modelling, more traditional decision theoretical techniques were used for word prediction \cite{vandenBosch2005ScalableCW}.} 

This means the training of the core model is unsupervised (or what is called `self-supervised' \cite{9010283}). Being able to do unsupervised training is an enormous advantage, because it enables automated collection of data in many different natural languages, programming languages, and even formal languages, without having to bother about what information these token strings actually contain. All this data is then used for training the above described target function in a time-consuming calculation using dedicated computing machinery costing enormous amounts of energy and money. The result is a model with billions of learned parameters that is not trained further, but used as the core of subsequent models, which is why it is called a `pre-trained' or `foundation' model.

\smallskip \noindent {\bf Fine-tuning}\\ 
An LLM, however big, will not spontaneously start to work as a conversational agent. A lot of functionality needs to be added. A chatbot needs to behave in a dialogue fashion; for instance, it should at some point stop producing more words and sentences as an answer to a question. Also it needs to know how to perform tasks the user asks it to perform (doing translations, rewriting in a certain style, finding the mistake in a program, etc.). This task-oriented behaviour is typically not in the data the core LLM is trained on (although some of it might be, if it is trained on Twitter, Reddit, etc., which do have examples of the behaviour required for dialogue interaction and task interpretation). 

The process of training the chatbot on data selected and produced by the system designers in order to induce useful functionality is called `fine-tuning'. The learning mechanism is (usually) not much different from the mechanism underlying the training of the initial LLM, but \emph{the difference is that the data is carefully selected to induce specific linguistic behaviours}. Fine-tuning may target only the parameters of specific layers of the deep learning network underlying the LLM; specific layers that are thought to be important for inducing certain functionality. But those details will not concern us in this article. What is important, is to realise that fine-tuning is a separate and supervised training phase and that finetuning is \emph{not} about curating the core dataset (a topic we address in section \ref{The ethical version of the disconnect}), but about \emph{adding} curated data to induce desired functionality.

\section{Behaviours}
\label{Behaviours}
The first thing to discuss in preparation for the central claims of this article, are chatbot behaviours. Since the conversational bots we are talking about are not connected to the physical world through any other actuators than the characters they can display on a computer screen, all their behaviour is of the \emph{discours} kind. We can distinguish several kinds of linguistic behaviour, and the differences will matter for the discussion to come. Discussing the behaviours well enable us to explain how it is possible LLMs behave as conversational \emph{agents}. 

Below I give a list of linguistic behaviours we may observe conversational bots to have. The list is not randomly ordered: it goes from behaviours that should be associated with more \emph{local} patterns in token sequences\footnote{Note that for a chatbot, all conversational behaviour comes necessarily in the form of token patterns.} to ones that should be associated to very \emph{global} patterns. The global patterns cannot be established before the local patterns are learned. Before discussing this in more detail, let me first give the list. 

\smallskip \noindent {\bf 1. Spelling and grammar}\\ 
Chatbots need to be able to produce sentences that are correctly spelled and according to the grammar of particular languages. So, they can be said to have a certain spelling and grammar \emph{behaviour}. The traditional view (that I mostly adhere to) is that patterns of grammar and spelling are mere conventions that need to be learned first before we can get meanings started. Without compliance to these basic patterns, the language produced by chatbots would not carry much meaning for human interpreters. So, it makes sense to see grammar and spelling patterns as \emph{enablers} for attributing content. 

\smallskip \noindent {\bf 2. Making conceptual sense}\\ 
Sentences that are correctly spelled and are according to the rules of grammar, do not thereby make sense. They can get conceptual structures wrong. They can make category mistakes. Sentences like "the colour of that tree leaf is slow" and "the flower smells purple" are correctly spelled, and grammatically correct, but mess with our concepts.\footnote{Humans are very good at appreciating and seeing meaning in poetry that `plays' with this aspect of conceptual sense making. LLMs are rather bad at poetry.} They make no sense, because they do not align with the way we conceptualise our world. We cannot really say they are true or false; conceptual structures and their associated language are (again) \emph{enablers} of structures higher up in this list (and truth-telling is two steps up). At this level of `making conceptual sense' it is no longer plausible to characterise the patterns only as mere \emph{form}; the (imperfect) isomorphism between the conceptual patterns in our texts and the way we carve up our world conceptually, could itself already be seen as providing content of a rudimentary kind: it tells us something about how we humans perceive and structure our world (and might, for instance, say little about how bats or potential extraterrestrial beings perceive and structure our world). 

\smallskip \noindent {\bf 3. Idioms and Style}\\ 
The language produced by chatbots can take on a certain style and can make use of certain idioms. I think idioms and style are still mostly aspects of form. But not quite. Choosing between saying things in one way or another, or choosing for a certain style are not entirely content neutral (they convey a certain message). But, I will not go into this any further. I just want to point out that idioms and style are reflected in patterns that live on top of grammar and spelling patterns and the conceptual patterns. 

\smallskip \noindent {\bf 4. Truth-telling}\\ 
One of the most central behaviours of chatbots, one that is discussed extensively by researchers and non-researchers alike, is truth-telling. I do not need to explain here in much detail what that is, as every language user has an intuitive understanding of it; it suffices to say it is about to what extent the sense making sentences (level 2) produced by a chatbot tell us things that we think are true. It is of course not the case that every LLM-sentence expresses a proposition that can be true or false, but we are now talking about the sentences that do.\footnote{In this article we will often just talk about `the truth of a sentence' when we mean `the truth of the proposition expressed by a sentence'. Because sentences, syntax, tokens and patterns are so central to the discussion in this article, it will be undoable to always be more careful about the distinction.} Note that, again, the patterns in texts that reflect a certain degree of truth can only live on top of the patterns we already discussed above: we cannot assess the truth of a sentence if it is not sufficiently grammatical or not understandable in terms of our shared sense-making conceptualisation of the world. To drive this message home even more clearly: where conceptual sense making rules out \emph{nonsensesical} token patterns like "the colour of tree leafs is most often slow" (with a very large number of grammatically correct words as an alternative for `slow'), truth-telling in addition rules out \emph{sensible} but \emph{false} token patterns like "the colour of tree leafs is most often orange" (with only other colours as an alternative for `orange'). 

\smallskip \noindent {\bf 5. Being logical}\\ 
If a text starts with a claim that `$P$' (with $P$ a proposition) and at the end of the text there is the claim that `not $P$', then the text is inconsistent. The textual patterns that are associated with avoiding inconsistency are the \emph{logical} structure of a text. These patterns, again, live on top of the patterns we have already discussed above.\footnote{One could dispute this, and argue that logic is a more basic pattern than truth. After all, logic not only regulates the relation between \emph{actual} truths (things like "the colour of tree leafs is most often green"), but also the relation between fictional or supposed truths. That is, logic is not \emph{about} actual truths but about meta level constraints (\emph{logical} truths, or validities) on inter-related \emph{possible} truths. However, nothing what I will claim in this article will hinge on this order. I do think, though, that in terms of LLM patterns, truth \emph{is} more basic. My reason for thinking so is that truth is an aspect of propositions, which concerns a fairly local aspect of texts, while logic is about relations \emph{between} propositions, which is a more global aspect that lives on top of the structure imposed by the truth of propositions (as I said, `$P$' can appear at the beginning of a text, while `not $P$' appears at the end). And in the order I propose here, more local textual patterns come before more global textual patterns.} The associated behaviour is the chatbot's \emph{logical} behaviour. Under the header of logical behaviour, I include making correct arguments and performing simple arithmetical tasks like addition, but also more complex mathematical tasks like solving equations, and even making (or correcting) programs and proofs.\footnote{As demonstrated by the well-known Curry-Howard correspondence, programs and proofs are isomorphic structures, which is exploited by functional and logic-based programming languages.} Whether all of these inclusions are fully justified will not be our concern here. 

\smallskip \noindent {\bf 6. Being rational}\\ 
Building upon all the patterns already mentioned, we can now look at the ones that are linked to the performance of chatbots as \emph{rational} agents. I will not give a precise definition of what it is to be rational, but it is clear that it is about maintaining a certain effective balance between beliefs, objectives and actions. This balance is again a constraint; a constraint maintained by (most) authors of texts. So, insofar conversational agents behave rational, it is again manifested as particular patterns in the texts they produce. We can think about things like staying on topic, or not going against self stated interests, like when the bot first claims to try to convince you of a certain point and then says something that goes very much against it. We can discuss if this level actually lives on top of the previous one or that the two levels should be merged into one. But, again, those matters will not particularly concern us here. 

\smallskip \noindent {\bf 7. Being social}\\ 
Social behaviour of chatbots is about the socially relevant regularities and patterns they follow in interaction with human users. I like to think these are patterns that again live on top of the ones already discussed, but the higher we come in this hierarchy, the more difficult it becomes to make such claims; certain aspects of layers start to interfere more heavily with underlying or superseding layers, so to say. For instance, one can maybe be social, to some extent, \emph{without} being rational. 

\smallskip \noindent {\bf 8. Being ethical}\\ 
The ethical decisions a chatbot makes are about \emph{what to say and what not to say} to avoid harm to others. This is a highly precarious and difficult task, as it can be hard to find out how not to harm others through texts.\footnote{I am thinking here about all the discussions on the bounderies of free speech.} Also, the way I portray it now, is a stark over-simplification, since not all texts that are experienced as harmful to some people should be considered unethical. However, I do think it is safe to assume that with ethical behaviour, we have reached the top level of the hierarchy.\footnote{I think that for texts, there is a good reason to think that the ethical patters live on top of the social patterns. Social norms for texts tend to be more local, being about how to avoid awkward formulations and about being polite and attentive to the readers possible concerns. Ethical textual behaviour seems to involve much more complex structures and difficult trade-offs (dilemma's).} 

\smallskip \noindent {\bf Learnability}\\ 
The first observation I want to make about the above hierarchy of behaviours is that because they are \emph{all} patterns in tokens/texts, they are in principle machine learnable.\footnote{In general, it is not the case that one machine learning algorithm performs well on every possible pattern (see the discusion on no-free-lunch theorems \cite{Sterkenburg2021}); good performance in one corner of the pattern space always needs to be paid for with bad performance in another corner. Note that in our case, the patterns build on each other, which makes it that we can also see them all together as one complex pattern. And that particular pattern turns out to be learnable quite efficiently with the transformer architecture \cite{Vaswani2017}. Why that is the case, we do not understand well. It is hard to believe it is a coincidence. But, why it is not a coincidence is not a question I will address here.} This explains why LLM-based chatbots actually work. No-one should be surprised that with enough training, machine learners start to follow all the concrete patterns exemplified to them by billions of text authors.\footnote{Some claim LLMs show `emergent' behaviour, since they have started to behave in ways not expected by their creators. I find it hard to interpret such claims as saying something significant. Since all linguistic behaviours are necessarily determining mere patterns in texts/tokens, nothing has `emerged' other than what could be expected with enough computation power.}However, patterns lower in the list are easier to learn than patterns higher up in the list. And this is because the patterns higher up in the list reside on ever higher levels of textual abstraction, that \emph{only} can be detected once the lower level patterns have been established. Actually, we find a clear example of this \emph{within} the first layer. I took spelling and grammar together in this layer, because the difference does not matter for my story. But I could have opted to split this level up into a lower spelling level and higher grammar level. Grammar is a clear example of a pattern that can \emph{only} be learned adequately if the more basic pattern of spelling is acquired first. 

\smallskip \noindent {\bf Classical AI}\\ 
To classical linguists and AI researchers, the above hierarchy must feel rather familiar, or maybe even old-fashioned. And that is because it is. The project of classical AI was to go through a very similar hierarchy step by step, in terms of explicit representation languages, explicit models, representation theories, rigorous proofs, and algorithms bringing it all to life. But it does not follow that modern AI is any different in how we should \emph{understand} what is going on. In LLMs, the same hierarchy helps us understand how every level can be learned as a pattern, and how higher level patterns require lower level patterns for their expression. The unique opportunity of LLM-based AI is that in its learning of the representations within the hierarchy, it targets the whole hierarchy \emph{at once}, by taking token prediction as its target function. No need for formal translations and proving their correctness, no need for representation theorems, no need for being worried about complexity classes, no need to devise algorithms to work with the representations, but just all of it in one go. And without the burden of formally linking all the levels.

\section{Knowledge}
\label{Knowledge}
To understand the central claim of this article, in section \ref{The knowledge-behavior disconnect}, about LLM-behaviour not following from declarative LLM knowledge, we first need to spend some time on the question of if, and how, we can talk about the knowledge of LLM-based chatbots at all.  

\smallskip \noindent {\bf LLMs and knowledge}\\ 
Many users easily talk about the information LLM-based chatbots give us, as indicative of the \emph{knowledge} they have. That is no surprise. Chatbots report us on how, according to them, things are, and do that declaratively, in languages we understand. And, if we set aside hallucinations for the moment, what they say is often in accordance with what we know to be true about the world. And if the facts of our world would have been different, the things being said by LLMs would also have been different. So, the things LLMs say \emph{are} a reflection of what is true about our world. The same point is exemplified by the patterns we discussed in section \ref{Behaviours}. These patterns depend on what is true about our world and on how we conceptualise it. This aligns well with what Dennett  \cite{Dennett1991-DENRP} says about patterns and AI. Dennett points to the fact that the patterns learned by an AI are (1) real (because not random, but determined by how the world \emph{is}), and (2) sufficiently meaningful to be regarded as knowledge by applying his intentional stance \cite{Dennett}. 

However, most philosophers would point out that since Plato, we have discussed specific necessary and sufficient conditions on human knowledge, and that in order to attribute knowledge to AI, it should maybe meet these same conditions. And then, if, as Plato proposed, knowledge is characterised as \emph{justified true belief} \cite{Gettier1963-GETIJT-7}, we can actually easily see reasons to claim that an LLM's answers \emph{cannot} be regarded as reflecting its knowledge. The statements an LLM makes are \emph{not} clearly externally justified (for instance through direct external causal links \cite{Goldman1967}) or internally justified (for instance, by the availability of reasons for the claims), are often \emph{not} true (the subject of many scathing take-downs of apparently serious answers given by LLMs), and it is hard to think of a way in which we can say that an LLM \emph{believes} them. I think believing is agentive in the sense that it involves the mental \emph{act} of committing to the truth of a proposition in the face of uncertainty about its epistemic status, and neither this act of `committing' nor the hesitation associated with `uncertainty' are easily attributable to LLMs. 

What this shows, I think, is that we should be hesitant to follow the route of applying well-known epistemological conditions and views to AI knowledge. Epistemology is about understanding \emph{human} knowledge. In understanding human knowledge, we do not have to worry about running the risk of explaining it in terms of agentive concepts that do not apply to the entities that are purported to posses that knowledge. But, with AI, we run that risk. We need to take a different approach.

\smallskip \noindent {\bf A pragmatic view}\\ 
I will suggest a more pragmatic view, one that is common in computer science and in natural language processing \cite{Petroni2019}. The only way in which we can interpret an LLM as an AI with which it makes sense to interact in a way that is comparable to how we interact with humans, is to say that it \emph{has} knowledge. Yes, it is often wrong. And yes, it is piggybacking on the knowledge in the texts it is trained on and has no independent ways to directly or indirectly verify or justify the knowledge, like we humans have. But what is most important for our purposes here is that we have to see the answers it gives as \emph{the knowledge} it has in order to assess its capabilities as a conversational AI. And that is the more \emph{pragmatic} notion of knowledge I will use in this article. 

It is actually in this same pragmatic way, that we talk about \emph{texts} containing knowledge. Making the comparison with textual knowledge explains the hesitation I expressed with applying the standard epistemological conditions to assess if what we are talking about is really knowledge; texts do not believe, or commit or hesitate in an agentive way. Their authors do, but that is not what we are discussing here. Note, furthermore, that this view extends to all levels of the discussed hierarchy. Individual texts do not only have a truth telling behaviour by which we assess their knowledge, but also have logical, social, ethical, etc. behaviour. It does not follow, of course, that texts are \emph{agents}. Again, only their authors are. 

The view on LLM epistemology I am advocating for here, has close connections to Popper's view laid out in his article "Epistemology without a knowing subject" \cite{Popper1968}. Popper distinguishes between `the second world' (the world of consciousness and knowing subjects) and the `third world' (the world of knowledge as it is reified in books, arguments, proofs, theories, etc.). Traditional epistemology belongs to the second world. But LLM-epistemology, I submit, belongs to Popper's third world.

\smallskip \noindent {\bf Knowledge-based systems}\\ 
In light of the above, it is instructive to know how knowledge is viewed in classical AI. In section \ref{Behaviours} I said that classical AI aimed to implement the behavioural hierarchy through explicit knowledge representation. And one of the core paradigms in linking the behavioural levels is that of the so called \emph{knowledge-based system}. The guiding idea of knowledge based systems \cite{hayes1983building,shortliffe2012computer} is that intelligent behaviour, in relation to a certain environment, is \emph{based on} knowledge about that environment. It follows that the intelligence of a knowledge-based agent depends on the correctness of that knowledge. And it follows that there is a mechanism that makes the behaviour counterfactually dependent on the knowledge: had the knowledge been different, the behaviour would (most likely) have been different too. So, a knowledge-based agent's knowledge about its environment \emph{feeds into} its choice of action for that environment. The knowledge is \emph{about} the environment and it contains --- in a for the agent accessible form --- information that enables the agent to make intelligent choices. And intelligent choices are typically defined in AI as choices that further the agent's interests, where those interests follow from the agent's goals (which are given to it by its programmers; let us not get distracted here by considering that agents might have or develop goals that are not given to them by their designers) \cite{Legg2007ACO}. The knowledge about the environment might have been collected during previous encounters with that environment, or it might have been sourced from other agents having had encounters with the environment. But, the crucial element of the idea of knowledge-based systems is that the knowledge is \emph{used as the basis for the decisions the agent makes}. And if that knowledge would have been different, the behaviour would most likely also have been different. 

The point I want to make about knowledge-based systems is that for such systems, behaviour and knowledge are \emph{necessarily} aligned. There is no room for non-alignment or, what I will call, in the next section, a disconnect, because a knowledge-based system's behaviour \emph{follows} from its knowledge. Knowledge-based systems are behaviourally, necessarily self-aligned.

\section{The knowledge-behavior disconnect}
\label{The knowledge-behavior disconnect}

%\smallskip \noindent [{\bf Declarative knowledge about behaviours}]\\ 
We will now compare the declarative knowledge LLMs appear to have (section \ref{Knowledge}) with their own behaviour (section \ref{Behaviours}). That is, we will compare the declarative knowledge \emph{about} that behaviour with the behaviour itself. And I will argue that the two are not always in line, and that when they \emph{are} in line, it is not because there is a mechanism in a chatbot that \emph{enforces} them to be inline. 

\smallskip \noindent {\bf An observation on truth-telling}\\
If we want to evaluate the knowledge we attribute to an LLM in relation to its behaviours, the truth-telling behaviour (the fourth level in the hierarchy of section \ref{Behaviours}) will obviously play a special role. The LLM's truth-telling behaviour is actually the behaviour \emph{by which} we are going to attribute declarative knowledge to the LLM. But, note that for this particular behaviour itself (i.e. for truth-telling), by definition the behaviour is in line with the content of that behaviour\footnote{Keep in mind that behaviour in this context is entirely textual.}; we cannot have that the truths being told might diverge from the behaviour, because the truth-telling \emph{is} the behaviour.\footnote{This is supported by the idea that for the particular behaviour of truth-telling, knowledge-how seems reducible to knowledge-that (knowing \emph{how} to tell the truth is reduced to something like knowing \emph{that} you tell the truth in cases where you chose to do so). But, making this claim more firm requires further analysis.} For the other behaviours (relating to grammar, logic, ethics, etc.), this is different: they will be measured \emph{against} the truth-telling behaviour to see if they are in line with it.

\smallskip \noindent {\bf An assumption about truth-telling}\\
However, there is a possible other problem with truth-telling behaviour. One might think that the claims being made by an LLM should not be taken at face value as the propositions it considers to be true. One might think, for instance, that LLMs can lie (say things they know/believe to be untrue). Or that they can give you bullshit (say things they do not know to be true \cite{Frankfurt1986-FRAOB}). For the moment, I am going to set aside such possibilities by making the following assumption. 

\begin{quote}{\bf Assumption 1}
The truth-telling behaviour of LLMs (level 4) can be taken at face value as indicative of the declarative knowledge they possess. 
\end{quote}

I think this is a reasonable assumption. I will not try to defend it in this section, but I will defend it against two possible attacks (one concerning the possibility that LLMs lie, and another suggesting that Davidson's challenge \cite{Davidson2007} to explain `radical interpretation' implies that LLMs maybe speak a different language, unknown to us).

\smallskip \noindent {\bf The central questions}\\ 
Let us now turn to the issues that are the central topic of this article: to what extent is the declarative knowledge LLMs report on in their truth-telling behaviour in line with their other behaviours? And what are the consequences for seeing LLMs as knowledge-based agents? I will try to state the central claims clearly. 

\begin{quote}{\bf Claim 1}
The conversational behaviours LLM-based chatbots enact are not \emph{informed} by the declarative knowledge they report on in the conversations you have with them. Their declarative knowledge and their behaviour is disconnected. 
\end{quote}

\begin{quote}{\bf Corollary of claim 1}
The declarative knowledge LLM-based chatbots report on in the conversations you have with them is never the \emph{basis} for the conversational behaviour they show in those conversations. In that sense, we cannot see them as (declarative) knowledge-based agents. 
\end{quote}

\smallskip \noindent {\bf Knowing the rules of Chess}\\ 
I will start by giving empirical support for claim 1. That there is a knowledge-behaviour disconnect comes to the fore very clearly in LLMs playing Chess. When we ask an LLM what the rules of Chess are, it will easily churn out all the allowed moves and general rules. It will also report on the slightly more obscure rules that many amateur Chess players that claim to know how to play Chess, are often not aware of. So when it comes to self-reporting on knowing how to play Chess, chatbots score 100\%. But let us now look at their behaviour. We can easily play Chess with such bots. They play Chess fairly well, and that is not surprising. In the texts they have been trained on, there are millions of Chess games. And them deciding on the next move in a game they play with you is just about choosing statistically optimal continuations relative to the games they have seen in their training set. 

But then, suddenly, without a warning, in a Chess game you play with an LLM, something strange happens: the system plays an illegal move (there have been many reports on illegal moves by Chess playing LLM-based bots). Of course that will surprise you, especially since it seems to play Chess fairly well. But an even stronger source of surprise is that previously, when you asked it about the rules of Chess, it only gave perfectly correct answers. So why does it not in all cases behave according to the \emph{knowledge it says to have} about the rules of Chess? Why does it not \emph{follow} the rules that constitute the game of Chess, while it reports to know these rules?\footnote{I believe this terminology to be in line with Wittgenstein's comments on rule following \cite{Wittgenstein_PI}. We can say that LLMs behave often \emph{in accordance} with rules, but that they have no way of actually \emph{following} them in the way humans can. In this article I will not pursue this connection further.} This is the disconnect I am talking about. 

\smallskip \noindent {\bf Knowing how to play Chess well}\\ 
We can extend the Chess example somewhat. We may argue that despite the occasional illegal moves, LLMs know how to play Chess \emph{well}. Also, they are able to give you quite a bit of information on Chess strategy. The LLM will give you information on openings, on end games, on tactics, etc., because there will have been a lot of Chess books in the data it is trained on. It is tempting then to think of the LLM as a knowledge-based agent, and to infer these two (playing well, and being to able to produce texts on strategy) should be in line \emph{because} the playing is guided by the information contained in the strategies the LLM reports on. But that is not how the mechanisms underlying LLMs work. None of the texts on strategy will be used as input for how an LLM plays Chess. For instance, an LLM might give you an opinion on whether some opening is better suited against a certain well-known player, because that was in the data it was trained on. But if you then go on to tell the system that you are that player, and you start a game of Chess with the system, there is no guarantee \emph{at all} that it will play that opening, because it will only play Chess on the basis of the games in the data it is trained on, but not on the basis of the strategies it reported on. It does not make the \emph{connection}. Thinking of it as a knowledge-based system, is a mistake.

\smallskip \noindent {\bf Performative contradictions}\\ 
The Chess behaviour just discussed, is an example of a performative contradiction. Performative contradictions were put forward in discourse theory \cite{Hintikka1962,Habermas1990}. They are situations where the performance of a speech act contradicts the contents of that same speech. A classical example is `I am asleep'. An example I particularly like is `it is needless to say that \ldots'. I see performative contradictions by LLMs as support for my claim that there is a disconnect of the kind I described. The Chess example is only one of many such possible examples. One small experiment I did was to make an LLM confirm that it is prudent for LLMs to comply with special requests of users, after which I asked it to stop responding to my prompts. And as you probably expect, it could not. It cannot stop talking, even after confirming to you that it \emph{should} and that it \emph{will} if that is what you want.\footnote{This example may seem unfair, as it seems to exploit the LLM feature of being hard coded to always give an answer. However, I think that the connection between declarative knowledge and behaviour should go both ways: if it cannot refrain from giving answers, it should be able to report that.} Another performative contradiction that is quite common is that it apologises after you point out that it has made a mistake, after which it continues to make exactly that same kind of mistake again. It apologises to you, not \emph{because} it made the mistake, but because it was fine-tuned to apologise whenever the user complains about it having made a mistake. It does not have a mechanism that aims at avoiding having to apologise, nor does it have a mechanism prompting it to take a stand on its belief in the truth of a proposition.

\smallskip \noindent {\bf Explaining hallucinations}\\ 
More empirical support for the disconnect comes from the abundance of hallucinations by LLMs. Hallucinations are textual behaviours. And as such they necessarily follow all the behavioural patterns we discussed in section \ref{Behaviours}. However, their \emph{content} is very much off. And I think that hallucinations are a symptom of the problem addressed in this article. The behaviour and the content are free to diverge, exactly because the content is \emph{disconnected} from the behaviour. And this typically happens if the content goes into uncharted territory, like counterfactual or counter-possible situations \cite{Zhang2023}. And once the content diverted away from familiar (that is non-counterfactual or non counter-possible) terrain, it will only head for even more unfamiliar terrain, because the knowledge contents of the linguistic expressions produced have no limiting impact whatsoever on the linguistic behaviour itself. 

In the literature \cite{Ji2023}, LLM hallucinations have been attributed to many different causes, such as, heuristic data collection, innate forms of divergence, imperfect forms of learning, erroneous decoding techniques, and different kinds of biases. I think all these analyses give only part of an answer, and point towards initial conditions that make hallucinations more likely. But, it seems to me that the root cause is that LLM behaviours are disconnected from their content. They simply keep following the patterns, however far the content of what they say has gone off the rails.

\smallskip \noindent {\bf Absence of a connection mechanism}\\ 
The support from performative contradictions and the abundance of hallucinations is empirical. A more fundamental argument for my claim follows from the limitations of the techniques underlying LLMs. As already explained the main technique is token prediction.\footnote{Underlying token prediction are non-recurrent deep neural nets called `transformers' \cite{Vaswani2017}. But other AI-techniques could in principle have been used. For instance it is not unthinkable to do token prediction using reinforcement learning or even classical decision trees \cite{vandenBosch2005ScalableCW}. It all comes down to scalability and performance, in which transformers excel for this application.} There are further mechanisms, and the most interesting one is the attention mechanism \cite{Bahdanau2014NeuralMT}, that learns to recognise more global patterns and can suitably be parallelised to efficiently exert influence on token prediction. Since attention mechanisms learn what wider contexts of sequences are relevant for their continuation, some would probably say they come close to an \emph{interpretation} of sequences linking their `content' to their continuation. But I think that is a stretch. Attention is still exclusively about the lowest level patterns. It is nothing like a traditional interpretation mechanism that has access to the linguistic content of sentences as we understand it. So, the question remains: can token prediction, as the core mechanism, ensure that the declarative knowledge contained in sentences, influences an LLM's linguistic behaviour?

I argue that with token prediction as the core mechanism it is possible to train \emph{maximally contradicted systems}. That is, based on the insights in section \ref{Behaviours} on how LLM-based chatbots work, we \emph{could} take the contradictory aspect to its extremes. And if the declarative knowledge an LLM reports on would be the \emph{basis} for its behaviour, like in a knowledge-based system, it should not be possible to train LLMs where the two go in maximally opposite directions, thereby showing the two are entirely independent. 

I did not actually train maximally contradicted systems, since I think our current systems already give us enough empirical support for the presence of performative contradictions, as I discussed above. But we can do some thought experiments. We can imagine training an LLM on individual texts that are already (maximally) contradicted in the way I described. Of course, these are strange texts. For instance, we could have texts on Chess that have the reporting on the rules of Chess very wrong and yet discuss and present moves in particular existing games that are according to the correct rules. We, as human language users, would see the problem immediately. We would not understand such texts, or, at least, would know which parts to ignore to understand them. But an LLM trained on these texts would do nothing more than learn the patterns, as that is \emph{all} the token prediction mechanism does. Then, after training, if it would be asked to play Chess, it would dutifully oblige, and if it would be asked to give the rules of Chess, it would give rules that squarely contradict its own playing. It would do it without a sweat, as those were the patterns it learned. It simply has no way to `see' that the texts it was trained on were in an important sense internally conflicted. There is nothing that signals anything contradictory to the machine, because there is no mechanism connecting knowledge content and behaviour. 

The same reasoning applies to almost every level discussed in section \ref{Behaviours}. For instance, for the first level, imagine that an LLM is trained on a very large body of English texts on English grammar that puts forward a grammar that is wrong. Maybe it puts forward a grammar of English that is internally consistent and has nice features, but that is simply not the grammar according to which the texts themselves are written. We can construct such texts, there is little doubt about that. They would be strange texts, clearly. If someone were to read the texts out loud, she would run the risk of being accused of performatively contradicting herself. And in that case, she would probably claim: "but, I am just reading what it says here, I would not want to endorse any of this!". However, an LLM trained on the texts would not have any qualms, as it has no mechanism to detect the conflict. It would start to behave just as inconsistently as the texts are.  

If we go to higher levels in the hierarchy of section \ref{Behaviours}, we can follow the same reasoning and make believable that we can train highly conflicted LLMs. And on the higher levels, the conflict can also be of a complementary kind, where the rules and principles are the correct ones, but the behaviour itself is very much off. This other (complementary) way of training a conflicted LLM is not possible for the first two levels (spelling, grammar and conceptual correctness) since if we cannot read and understand the output of an LLM, we cannot get the truth telling behaviour going and cannot attribute knowledge to the LLM at all. Therefore, one should not think that disconnects on levels lower than truth telling are a threat to the assumption that we can take what they say at face value as the declarative knowledge they possess (assumption 1). 

Let me demonstrate this for the highest level mentioned in section \ref{Behaviours}, the level concerning a chatbot's ethical behaviour. We could train the LLM exclusively on brutal texts on discourse ethics that are offensive to the reader, but that contain ethical theories and rules that go directly against those textual behaviours and actually preach prudent discourse practices. An LLM-based chatbot trained on these texts would not have a mechanism to detect the problem and would not practice what it preaches. The absence of a connection between the two would prevent that. 

It seems clear then that we can tell this story about training highly conflicted LLMs for every level discussed in section \ref{Behaviours}, with the caveat that for some levels we need to be careful about how to do that exactly (not for any level we can apply both complementary ways to get a conflict). But, there is one level for which the story does not go through: the truth-telling level. This links directly with the observation this section started with, the observation that for truth-telling knowledge and behaviour are \emph{necessarily} in line with each other. It is instructive though, to see what goes wrong if we \emph{try} to tell the story about training an LLM that is maximally conflicted when it comes to truth telling. One of the two complementary ways to train a truth conflicted system, is to train on texts that give the \emph{wrong} instruction on what it is to tell the truth, while at the same time, what they do, is tell the truth. The point here is that texts that "give the wrong instruction on what it is to tell the truth" and at the same time "tell the truth" are impossible.\footnote{Had we tried to tell the story with the complementary way to come to a conflict, the result would have been the same.} Such texts cannot exist.\footnote{Very much like the barber who shaves exactly those who do not shave themselves, cannot exist \cite{Russell1918}.} So, for truth-telling behaviour we cannot get the argument going. Truth-telling is a special linguistic behaviour for which there cannot be a disconnect, because the things we think to distinguish in our analysis (behaviour and declarative knowledge), happen to be just the same thing.\footnote{I think we cannot train a maximally truth conflicted LLM by training exclusively on liar sentences \cite{Barwise1987}. The point about liar sentences is that we hesitate about their truth (the truth of the propositions they express). But that is not what we are aiming for in this argument; we are aiming for an LLM that produces sentences that \emph{are true}, while what they say amounts to \emph{incorrect guidance} about what it is to tell the truth. Liar sentences are not like that. We are not sure \emph{if} they are true (express true propositions). And what they say, \emph{is not} guidance about what it is to tell the truth. And therefore liar sentences can exist (and do so). But, the maximally truth conflicted LLMs we aim at here, cannot exist.}

\smallskip \noindent {\bf Fundamentality}\\ 
If I am correct, the disconnect is a fundamental consequence of the core technique of token prediction that underlies LLM-based chatbots. It is there at the lowest level of grammatical behaviour and it is there at the highest level of ethical behaviour. And it is independent of the training effort conducted for an LLM. Of course, the more we train, the more the disconnect will be obscured from sight. And to a large extent, in our current systems, it already is. For instance, for knowledge of grammar and spelling of the English language, our current systems do not seem to be disconnected at all: asked about those grammar rules and the spelling of words, a conversational bot will give perfect answers which come in a form that actually is perfectly  \emph{in accordance} with those exact same rules. So no symptoms of a disconnect. But as the possibility to train highly conflicted systems shows, a disconnect \emph{must} be there. The only reason that it does not \emph{show}, is that it is sufficiently trained on texts by humans, who do \emph{not} suffer such a disconnect and who produced texts that are, on average, \emph{not} internally conflicted. 

We can phrase this in terms of the difference between correlation and causation. We may say that declarative knowledge and behaviour \emph{correlate} in LLMs, but without LLMs having a mechanism that is responsible for the first being a \emph{cause}\footnote{One might wonder if it is correct to talk of `causes' in this context. But, I think it is. An LLM pattern containing information that is interpretable as declarative knowledge, is a physical structure, and could in principle, through some mechanism, \emph{cause} the right behaviour. The point is that such a mechanism is absent.} for the second. There is a \emph{common} cause that originates in the human authors of the texts the LLM is trained on. We should not make the mistake to project this human causal mechanism into LLM-based chatbots only because we observe an (imperfect) correlation. The correlation can be made stronger by more training on more data, but it does not follow that the correlation will spontaneously turn into a causal connection at some point. This supports the following claim. 

\begin{quote}{\bf Claim 2}
The (declarative) knowledge - behaviour disconnect is a fundamental limitation of token prediction-based chatbots; it does not dissolve with more training, even though it might look as if it does. 
\end{quote}

\smallskip \noindent {\bf Saliency at higher levels of behaviour}\\ 
Disconnects will be more and more obscured from visibility by more intensive training on more extensive data sets. But, as I already pointed out in section \ref{Behaviours}, the higher the level of behaviour, the more difficult it will be to learn the associated patterns. This means that we should see the most of a disconnect with these harder to learn patterns. The reason is that it only takes lower level, more easy to learn patterns to be able to produce \emph{declarative knowledge} about the higher level social or moral behaviours, while the higher abstraction and lower learnability of these higher level behavioural patterns will increase the likelihood of them not aligning with this knowledge. 

\begin{quote}{\bf Claim 3}
The higher the level of behaviour, the more the disconnect comes to the fore.  
\end{quote}

\section{LLM alignment}
\label{The ethical version of the disconnect}

As I see it, and as I explained in section \ref{Behaviours}, ethical behavioural patterns build upon all the other behavioural patterns (for truth, logic, rationality, sociality, etc.). So, ethical linguistic behaviours are among the most abstract textual patterns to learn for LLMs. And because they are the most abstract patterns, they are the hardest patterns to learn (since first lower level patterns need to be learned).

\begin{quote}{\bf Corollary of claim 3}
The disconnect comes to the fore most prominently in ethical LLM behaviour.  
\end{quote}

And, I think that this is exactly what we see in the chatbots that are currently on the market, which gives further empirical support for the views put forward in the previous sections. Ethical considerations are everywhere; in political texts, in philosophical texts, in newspapers, in social media posts, wherever you look. And because chatbots are trained on \emph{all} these texts, you can have fairly good conversations on ethical issues with them. However, if I am right about the disconnect, all this information is \emph{not} used as the basis for the chatbot's own ethical behaviour. And I think we can see that is true, because the biggest concern we have had with chatbots is their refusal to behave ethically. Despite all the books they have `read' on ethics, politics, law, and raising children, they have no clue about behaving ethically. And because behaving ethically is difficult for LLMs, designers have come up with alternative techniques to make them behave better. I will briefly discuss three of them.  

\smallskip \noindent {\bf Alignment by curating training data}\\ 
The first thought AI designers have if they want to better align a deep learning model that displays inappropriate behaviour, is to curate the data (deciding beforehand what to train on and what not, for instance, to avoid bias or prevent harmful behaviour). So, we can go from an unsupervised set-up to a more supervised one. However, this will do nothing to solve or mitigate the disconnect. As a consequence of the disconnect, we actually face a choice: do we curate texts on their unethical textual behaviours\footnote{As I argued in section \ref{Knowledge}, it makes sense to talk of the `behaviour' of a text in exactly the same way as it makes sense to talk about the behaviour of an LLM.} or on their unethical textual content? Probably we will want texts that perform well on both criteria. But that, again, is only a way of \emph{hiding} the disconnect.

However, this problem is not the only reason that curation is not a main technique for aligning LLM-based chatbots. The problem is that curation defeats the brute force learning approach to LLMs. It obviously stands in the way of the idea that to get to a conversational agent, we need to train on \emph{anything} textual we can get our hands on while \emph{not} having to bother about selecting what is appropriate and what not. The sheer size of the data of such an approach simply makes curation practically infeasible. 

\smallskip \noindent {\bf Alignment through reinforcement learning}\\ 
Since curation of data is not a good option, programmers came up with a special form of LLM finetuning that makes use of reinforcement learning. The idea is to first train a \emph{reward} model that \emph{values} outputs generated from specific inputs. The rewards are sourced from human feedback in a reinforcement learning feedback loop that goes through an iterative process of trial and error. In a second step, this reward model is then used to finetune the LLM by employing the same deep learning mechanism (the transformer) that was employed to train the core.\footnote{So, it is not the case that a pre-trained LLM is seen as the initial policy to be fed into a policy iteration algorithm as commonly used in reinforcement learning, as one might think. I would not be sure if such an approach would be practically feasible.} This way, an LLM can be directed towards desired behaviours by giving the right human feedback in training the reward model (that in turn fine tunes the LLM). The approach is called `reinforcement learning through human feedback' (RLHF) \cite{Ouyang2024} and has become a popular way to align LLMs. 

However, the technique comes with some serious problems. Because the behaviours aimed at (i.e., ethical behaviours), are at higher levels of abstraction, the finetuning on the basis of the reward model is typically only performed on the final layers of the network. Or, alternatively, the reward model is used to add an extra, filter-like layer that selects between LLM outcomes (text continuations) that are close alternatives. This results in a situation where the main LLM model at the core still contains a lot of unaligned information. Since LLMs are trained on anything the designers could get their hands on, they will have inherited some ugly conversational dynamics from social media and dubious internet forums (like 4chan). RLHF then adds behavioural corrections to the system by curbing certain unethical outputs. But as so-called `jailbreaks' have shown, it can be fooled to release the bad content at its heart anyway. 

Apart from the jailbreaking issue, there are more problems. I think by using RLHF, an even further disconnect will arise. The human annotators performing the supervision may of course be capable to articulate why, and according to which moral principles, they penalise certain (types of) outputs, but these articulations are \emph{not} added as data to the core LLM that produces the outputs in the first place. So, the ethical behaviour learned by the system through the RLHF technique is \emph{not} based on the ethical knowledge it may report on in conversations; the two simply come from \emph{different sources}. The ethical \emph{behaviour} is sourced from the supervision by human annotators, and the ethical \emph{knowledge} is sourced from the texts in the training data. And never is there any information exchange or interaction between the two. This clearly only aggravates the disconnect.

\smallskip \noindent {\bf Alignment through prompt engineering}\\ 
Prompt engineering is about \emph{instructing} the chatbot to behave in a certain way. It \emph{primes} the interaction with an LLM, often without the enduser being aware of it (which is why I suggest the term `shadow prompting' for this). This provides a third way in which we can try to align LLMs; we can simply \emph{tell} them, by engineering suitable prompts, to behave in the way we want them to behave. Well-known instances of prompt engineering are jailbreaking (providing prompts that tell an LLM not to behave in the prudent ways it was learned to do using RLHF techniques), and chain of thought prompting \cite{Wei2024} (giving the LLM step by step guidance on how to walk through certain problems). 

Also this third technique to implement ethical behaviour does not provide a mechanism that connects declarative knowledge to behaviour. The way an LLM reacts to a prompt, be it a user prompt or a shadow prompt, is determined by the continuation patterns it learned (section \ref{Behaviours}). This `behaviour' is what we want to connect, to what it is that is being said. It might of course be that if thought prompt engineering we \emph{tell} an LLM to behave in a certain prudent way (alignment) or non-prudent way (jail breaking), it will do so. But, it will only do so by virtue of the patterns it has learned. In particular, it will only do so on the basis of the \emph{social} and \emph{ethical} patterns it has learned (section \ref{Behaviours}), because these tell the LLM how to react to requests or commands. But see how strange this situation is: an LLM will only `obey' the ethical and social instructions it is given in shadow prompts to the extent that it \emph{already} has learned the social and ethical patterns that make it answer prompts politely and obediently in the first place. This is a strange situation. The ethical proficiency of an LLM should not depend on how well it is trained to slavishly obey the ethical instructions it is given through prompt engineering. And what is more, now there are even three things to align: (1) declarative knowledge picked up from the training texts, (2) declarative knowledge/instructions from the shadow prompts, and (3) the linguistic behaviour that is now based on both the behaviour that is learned from the behaviour in the training texts \emph{and} the behaviour induced by obedience to the shadow prompt instructions.\footnote{So, hidden in this structure is actually a mechanism where behaviour \emph{is} based on declarative knowledge, namely, the declarative knowledge in shadow prompts. However, this is at the expense of introducing other disconnects.} The possibility for disconnects has only increased.

\section{Objections}
\label{Objections}

I will discuss three different possible objections to what I have put forward. The first two objections are an attack on assumption 1 about taking the knowledge expressed in the outputs of an LLM at face value as the declarative knowledge the LLM possesses. The third objection is more general, and suggests that disconnects are no problem to begin with, because we humans also often show signs of being disconnected in the way described in this article. 

\smallskip \noindent {\bf LLMs might be lying}\\ 
A first objection against assumption 1 about taking the truth-telling behaviour of LLMs as indicative of the declarative knowledge they possess, is that we might think that LLMs can lie. Because, if they can lie, maybe they are just fooling us. Maybe their behaviour \emph{is} in line with the propositions they \emph{actually} consider to be true, which would mean there is no disconnect. But I think this is far fetched. I think LLMs cannot lie. Why would we assume they can? Should we assume they maybe can have learned to lie from being trained on texts by lying authors? I do not see how that could make sense in terms of the token patterns learned (section \ref{Behaviours}); there is no difference between token patterns that reflect \emph{lies} of text-authors and token patterns that reflect mere \emph{mistakes} made by these same authors. An LLM has no way to distinguish between these two (and one reason for that is that it has no independent way of verifying truth), but the difference matters greatly here. I think we should be careful in projecting human qualities (if lying is a quality) onto LLMs too quickly. Lying -- like believing -- requires a mental action. It requires (1) the belief in a proposition, and (2) a public commitment to the truth of the negation of that proposition.\footnote{Bullshitting \cite{Frankfurt1986-FRAOB} is slightly different; it requires (1) the \emph{absence} of an epistemic position regarding a proposition, and (2) a public commitment to the truth of that proposition.} There is no reason to assume LLMs can obtain the qualities required to entertain such attitudes.

\smallskip \noindent {\bf LLMs may need radical interpretation}\\ 
A second way to attack assumption 1 is to suggest that what is wrong with it is not that LLMs may hide the truth (the above objection) but that we may not understand correctly what they say. We might think that the knowledge expressed in the language that an LLM puts out to us, is not the knowledge it \emph{actually} has, because we may \emph{interpret} the language wrongly. If there is another \emph{interpretation} of the sentences an LLM produces, then these interpretations may give rise to different attributions of knowledge to an LLM, ones that \emph{might} be in line with its behaviour.

This suggestion basically asks us to be more charitable: LLMs would behave in accordance with their knowledge if only we would not misinterpret what they say. Thoughts like these are reminiscent of Davidson's theory on radical interpretation  \cite{Davidson2007}. Davidson asks us to think about if it is possible to interpret sentences in a language unfamiliar to us (think about ancient inscriptions or signals from outer space) without having information on the beliefs of their authors, and without having prior knowledge of their meaning. Radical interpretation poses what we might call a `bootstrapping' problem: to discover how to interpret unfamiliar linguistic expressions, we need access to the beliefs the producer of the expressions must have had, but the only way to get this access is through the meaning of the expressions. Davidson suggests to solve this stalemate by starting to graciously attribute some beliefs first, and work out the meaning of the linguistic expressions from them, in a co-inductive interpretation building process. 

Does Davidson's principle of charity then also help us to solve the disconnect in LLMs? We can try something like the following. If the outputs of LLMs contain knowledge that is different from the knowledge pointed to by the standard (let us say English) interpretation, maybe we should regard these outputs as written in a language that is unfamiliar to us, a language that requires radical interpretation. And, following Davidson, we could start to find out what this interpretation is by charitably attributing beliefs to the LLM that are in line with its \emph{behaviour}, and work our way up in a co-inductive way to find out that there is actually no disconnect but only misinterpretation. But, I think at this point this line of thought goes wrong. If there was a different interpretation of the linguistic expressions on the basis of which we are going to attribute different beliefs to an LLM, also the linguistic \emph{behaviour} needs to be characterised in terms of it. It seems like we are dealing with a moving target then: the interpretation of the expressions produced is the basis for \emph{both} the declarative knowledge attribution \emph{and} the linguistic behaviour. So, we cannot assume there is a non-standard interpretation that only affects one of them and brings knowledge and behaviour in line. This line of reasoning leads us nowhere. 

I also see a second reason not to be charitable in the way Davidson has in mind. Our concern with the disconnect is one about rationality. Behaviour that is not in line with knowledge, is arguably an instance of irrational behaviour. But, one of the \emph{conditions} Davidson gives for applying his procedure is that the author of an expression in need of radical interpretation, is sufficiently like us, and, in particular, a \emph{rational} being. So, if we then want to apply Davidson's procedure to explain away the disconnect in LLMs, we run into a second, more problematic bootstrapping problem. We want to \emph{show} that aan LLM's behaviour is rational and in accordance with its declarative knowledge. If, by applying Davidson, we \emph{start} by attributing rationality, we are begging the question. 

Clearly, there is also independent reason to believe that the only sensible interpretations of linguistic expressions produced by LLMs are the standard ones. If it would make sense to assume that outputs contained knowledge that was only available to us by applying a non-standard interpretation of the language used, we would have to assume the same possibility for the texts the LLM is trained on. This, we never do. And for good reasons. We know texts are written by people like us, with mostly similar beliefs. So, there is no reason to suspect an alternative and therefore radical interpretation of LLMs: the knowledge contained in them is in the same way accessible to as the knowledge captured in the texts they are trained on. 

\smallskip \noindent {\bf Humans are disconnected too}\\ 
An obvious third objection against the claims this article puts forward is that disconnects of the kind I accuse LLM-based chatbots of having, also appear between \emph{human} behaviour and knowledge. People do not always \emph{do} what they \emph{know} to be the right or optimal thing to do. This then seems to suggest that we are on the wrong track here in thinking there is anything wrong with the presence of disconnects in chatbots at all: humans are disconnected too.

But, I think they are not. If an outside observer confronts a human with a performative contradiction she demonstrated, she will respond by giving reasons or excuses for it, and will likely start to reflect on the situation. As the result of this, the human agent will either amend her behaviour or her (self)knowledge, and unity will be restored. This process is highly important, as it makes human behaviour more adaptable and actually capable of (moral) progress. It also feeds into the idea that humans are responsible for what they do in relation to what they believe. None of this resembles in any way how LLM-based chatbots deal with this type of dissonance. There are simply no mechanisms in place for these bots to perform the consistency restoring activities I just described. An LLM \emph{will} apologise in situations like these, but not because it is aware of the dissonance; it only does so because it is trained to react to your complaint about its behaviour in that way. Humans, however, \emph{base} their actions on epistemic (and other) reasons, as can be seen from the consistency preserving operations they perform on them. And that is also how traditional knowledge-based agents in AI do it (using belief revision \cite{Hansson03}). A challenge is to bring this capacity to LLM-based conversational agents. And I think it is not a challenge that can be easily met with current ideas. 

So, my reply emphasises the role of the mechanisms at work. Of course, if you are a behaviourist, you do not care about the mechanisms; all you care about is what they amount to: the behaviour. It is unlikely then, that I have convinced behaviourists who just point to the fact that humans often show signs of disconnects too, who, furthermore, see how behaviour and content get more in line by ever more training, who see how LLMs keep improving, and who see how hallucinations get less with each new iteration of a chatbot. So be it. I think the mechanisms matter, because in important cases we are \emph{aware} of these mechanisms functioning in ourselves. And this fact has consequences for how we look at others. For instance, in situations where we see a human make a performative contradiction, we understand that something is wrong. We will never accept it as just a behaviour. We will either conclude the person is not aware of what it is doing, or maybe, that we misunderstood the message, or we come up with some other explanation to make the contradiction go away. More important, if we think the person that directed this speech at us is intelligent, we are sure we can explain to her that the performance \emph{is} contradictory, and that we will be able to convince her to either adopt her (possibly moral) stance, or her behaviour. Never will we assume the person will defend herself by claiming that what she did is \emph{actually} intelligible. And this is \emph{because} we know how the mechanisms work, from experiencing them at work in \emph{ourselves}. All that is lost on the behaviourist who is happy to look at the behaviour only, and is not missing anything in an LLM, as long as it behaves reasonably well in most situations.

\section{Conclusions}
\label{Conclusions}

This article has argued that the techniques underlying LLMs prevent LLM behaviour from being based on the declarative knowledge these same LLMs report on in conversations you have with them. The lack of dependency between the two was called a `disconnect'. 

\smallskip \noindent {\bf Knowledge and ability}\\ 
Many ensuing and otherwise related questions had to be left untouched. One is that the connection that needs to be established to get rid of the disconnect should run both ways. It is not only that behaviour should follow knowledge, also knowledge should stand in connection to behaviour. In particular, truly intelligent agents should have knowledge \emph{about} their abilities\footnote{The epistemology of ability is a relatively underdeveloped area in philosophy.} and \emph{about} their actions; knowledge they should be able to report on in the conversations you have with them. Clearly you can ask a system like ChatGPT what it is able to do. But the answer you get is one that is being fed to it by its programmers in the fine-tuning phase of its training trajectory\footnote{This is most obviously true for the first version of ChatGPT released to the wider public in 2022, that repeatedly claims it is trained on texts from before 2021, which means it cannot have learned about its own capacities through these texts.}; it has no access through introspection or through observation of its own actions. That is a mechanism it lacks. And again, that fact can be hidden from sight -- thereby fooling the behaviourist -- if its programmers, during fine-tuning, happen to learn it exactly the right information about its capacities (which is, of course, unlikely). One could speculate then, about how a mechanism relating knowledge and behaviour could look like. And it seems such a mechanism should provide some form of introspection that links awareness of linguistic behaviours to the knowledge contained in the texts produced. I do not know how such a mechanism could be constructed, and if it is possible at all, but current LLMs do not have it.

\smallskip \noindent {\bf Meaning and use}\\
A second closely related topic we have mostly avoided in this article is how to look at the \emph{meaning} of the sentences uttered by an LLM-based chatbot. We \emph{did} look at it briefly, in section \ref{Objections}, when we discussed Davidson's ideas on radical interpretation. But, that was in the service of being (too) charitable by attributing \emph{knowledge} to LLMs other than the \emph{knowledge} they report on in terms of the standard interpretation of the language they use; a suggestion that I rejected. Actually, in this article I have tried to carefully avoid talking about meaning. Our primary concern was with knowledge and not with meaning. And in section \ref{Knowledge}, where we discussed the relevance of the views of Dennett and Popper for LLM epistemology, I think we came to a relatively clear picture about what it means to talk about the \emph{knowledge} an AI has about the world. I find it much harder to get a clear picture of attributing meaning to LLMs. If we talk about meaning, we always need to keep in mind \emph{whose} meaning we are talking about: the meaning \emph{we} read into language \emph{or} the meaning the AI reads into language. And that second thought, `the meaning an AI reads into language' I find it very hard to make sense of. 

Yet, it is rather tempting to think that the claim made in this article has implications for the claim that meaning is nothing but use (generally attributed to Wittgenstein \cite{Wittgenstein_PI}). After all, the underlying LLM mechanism of token prediction trains explicitly on language \emph{use}, independent of its content. And then, the existence of the disconnect I have argued for, might suggest that this mechanism is \emph{not enough} to get to the content.\footnote{Wittgenstein \cite{Wittgenstein_PI} actually left open the possibility that meaning is \emph{not} exhausted by use. PI 43: ``For a \emph{large} class of cases of the employment of the word `meaning' -- though not for all -- this word can be explained in this way: the meaning of a word is its use in the language''} However, this is all I want to say about meaning and LLMs. A further investigation of meaning in LLMs, in relation to the claim that it is nothing but use, will be conducted in another article.\footnote{My claim in that article will be that meaning is indeed \emph{more} than use, and that LLMs present us with a unique opportunity to demonstrate that, by showing how they treat \emph{semantic paradoxes}.}

\smallskip \noindent {\bf Ryle and knowing how}\\ 
We focussed on the declarative knowledge LLMs have and how it is disconnected from their behaviour. But we did not discuss the possibility to look at what an LLM says, not as declarative knowledge, but as practical knowledge  \cite{Ryle1946}: practical knowledge of \emph{how} to answer to certain prompts. It is tempting to cast the disconnect in these terms: LLMs \emph{know} what to say, but do not \emph{know} what they say. That sounds right, because `knowing what to say' is about linguistic behaviour and `knowing what you say' is about declarative knowledge. Of course, we always think of the first (knowing what to say) as being constrained by the second (knowing what it is you are saying), but the disconnect alleges that for LLMs this consistency limitation is absent. 

It is also tempting to connect the disconnect to Ryle's famous claim that knowledge-how (practical knowledge) is not reducible to knowledge-that (declarative knowledge).\footnote{But, practical knowledge of the \emph{linguistic} kind, that is, knowledge of \emph{how} to react to linguistic prompts, is probably not what Ryle had in mind when he was thinking about practical knowledge.} And then it may even make sense to think that \emph{linguistic} practical knowledge is just (knowledge of) meaning, and to claim that linguistic practical knowledge not being reducible to declarative knowledge has a parallel in meaning-as-use \emph{not} being enough for meaning-as-content (or `sense', in Frege's terms \cite{Frege1948}). But clearly, all these suggestions need further investigation. 

Let me nevertheless spend a few more words on how attractive the picture is. The idea of an LLM, in reaction to a prompt, knowing what words to say (in the performative sense), without knowing what it is that it is saying (in the declarative sense) is reminiscent of the freshman philosophy student who does not know what a text says, but who can reproduce the fancy words and phrases to her teacher. In pedagogy, at least in the Dutch teachings on it, there is a word for this type of verbal behaviour: `verbalism'. Verbalism is the phenomenon of students using words because they (roughly) know their use is correct in the given context, but without knowing their actual content.\footnote{The English word `parroting', also alluded to in the title of \cite{Bender2021}, does not capture the same sense, because if one parrots a certain opinion, it does not follow that one does not know what it is one is saying.} Verbalism can be observed in children in elementary school, when the teacher asks them to explain a topic they have not understood. But, the phenomenon is not restricted to natural language. It can also be observed in high school students needing to pass their mathematics tests: many of them have little idea about what it is they are doing, yet they pass their math test purely by applying pattern matching and symbol manipulation.\footnote{There are clear links here with Searle's Chinese room argument and with a well-known position in the philosophy of mathematics. But I will not pursue these links here.} 

%Verbalism is sometimes also seen with students of (not seldomly continental) philosophy, who did not fully grasp the meaning of words, but know when to use them to impress their professor or corrector. 

\smallskip \noindent {\bf Implications}\\ 
Before rounding up, I would like to point to one possible implication of the theory put forward in the article. It has been argued that AIs should be explainable in terms of the \emph{reasons} for their actions \cite{Baum2022}. I agree that this is a commendable aim. But our results in this article show that the route towards it might be hard. Just asking an LLM to give reasons for why it says certain things, \emph{will} provoke it to give answers. But if the points I make are correct, those answers will not be \emph{based} on what it \emph{says} to know about why it gave those answers. Therefore, it will be very difficult to put trust in anything it might say about why it behaved in a certain way. 

\smallskip \noindent {\bf Linguistic behaviourism}\\ 
Let me, by way of finally concluding this article, try to reframe its main message in yet other terms. Most human language users select their words carefully. They have in mind a certain message that they want to convey to a certain public and choose the words that best fit that aim. It is what every writer does, including me now. This results in certain linguistic behaviours. I think there are some people who think that somehow this process can also be reversed; that by pooling word selection behaviours of billions of human writers together in a statistical way, we create a genuinely new word selecting agent. We may call this idea `linguistic behaviourism', since what that view is based on, is that mimicking or simulating linguistic behaviour is \emph{sufficient} to obtain a genuine agent who uses language to express what it knows and wants to convey. One way to look at this article is then to say that it claims that token prediction is a technique based on the idea of linguistic behaviourism, but that linguistic behaviourism is false. LLM linguistic behaviour is disconnected from the declarative knowledge expressed by the linguistic items that make up the behaviour. Our linguistic behaviour is not. 

%\section*{Acknowledgement}
%Thanks to TBD, TBD, TBD, TBD and TBD, the students of the course 'Philosophy for AI' at TBD, and the attendees of the the workshops TBD and TBD, for giving valuable feedback on (previous versions of) the material in this article.  

%John Jules Meyer, Davide Grossi, Clayton Peterson, Janneke van Dis, Joris Graff, Brandt van der Gaast, Niels van Miltenburg, Hein Duijf. 

%\printbibliography. % this is for the Biblatex/biber way of doing references.

\bibliographystyle{plain}
\bibliography{references}

\begin{thebibliography}{10}

\bibitem{Bahdanau2014NeuralMT}
Dzmitry Bahdanau, Kyunghyun Cho, and Yoshua Bengio.
\newblock Neural machine translation by jointly learning to align and
  translate.
\newblock {\em CoRR}, abs/1409.0473, 2014.

\bibitem{Barwise1987}
Jon Barwise and John Etchemendy.
\newblock {\em The Liar: An Essay on Truth and Circularity}.
\newblock Oxford University Press, Oxford, England and New York, NY, USA, 1987.

\bibitem{Baum2022}
Kevin Baum, Susanne Mantel, Eva Schmidt, and Timo Speith.
\newblock From responsibility to reason-giving explainable artificial
  intelligence.
\newblock {\em Philosophy \& Technology}, 35(1):12, 2022.

\bibitem{Bender2021}
Emily~M. Bender, Timnit Gebru, Angelina McMillan-Major, and Shmargaret
  Shmitchell.
\newblock On the dangers of stochastic parrots: Can language models be too big?
\newblock In {\em Proceedings of the 2021 ACM Conference on Fairness,
  Accountability, and Transparency}, FAccT '21, pages 610--623, New York, NY,
  USA, 2021. Association for Computing Machinery.

\bibitem{Davidson2007}
Donald Davidson.
\newblock Radical interpretation.
\newblock {\em Dialectica}, 27:313 -- 328, 05 2007.

\bibitem{Dennett1991-DENRP}
Daniel~C. Dennett.
\newblock Real patterns.
\newblock {\em Journal of Philosophy}, 88(1):27--51, 1991.

\bibitem{Dennett}
D.C. Dennett.
\newblock {\em The intentional stance}.
\newblock The MIT Press, 1987.

\bibitem{Frankfurt1986-FRAOB}
Harry~G. Frankfurt.
\newblock {\em On Bullshit}.
\newblock Princeton University Press, Princeton, NJ, 1986.

\bibitem{Frege1948}
Gottlob Frege.
\newblock Sense and reference.
\newblock {\em The Philosophical Review}, 57(3):209--230, 1948.

\bibitem{Gettier1963-GETIJT-7}
Edmund Gettier.
\newblock Is justified true belief knowledge?
\newblock {\em Analysis}, 23(6):121--123, 1963.

\bibitem{Goldman1967}
Alvin~I. Goldman.
\newblock A causal theory of knowing.
\newblock {\em The Journal of Philosophy}, 64(12):357--372, 1967.

\bibitem{Habermas1990}
Jurgen Habermas.
\newblock Discourse ethics: Notes on a program of philosophical justification.
\newblock In C.~Lenhardt J.~Habermas and S.W. Nicholsen, editors, {\em Moral
  Consciousness and Communicative Action}. Cambridge, Massachusetts: MIT
  Press., 1990.

\bibitem{Hansson03}
S.O. Hansson.
\newblock Ten philosophical problems in belief revision.
\newblock {\em Journal of Logic and Computation}, 13:37--49, 2003.

\bibitem{hayes1983building}
Frederick Hayes-Roth, Donald~A Waterman, and Douglas~B Lenat.
\newblock {\em Building expert systems}.
\newblock Addison-Wesley Longman Publishing Co., Inc., 1983.

\bibitem{Hintikka1962}
Jaakko Hintikka.
\newblock Cogito, ergo sum: Inference or performance?
\newblock {\em The Philosophical Review}, 71(1):3--32, 1962.

\bibitem{Ji2023}
Ziwei Ji, Nayeon Lee, Rita Frieske, Tiezheng Yu, Dan Su, Yan Xu, Etsuko Ishii,
  Ye~Jin Bang, Andrea Madotto, and Pascale Fung.
\newblock Survey of hallucination in natural language generation.
\newblock {\em ACM Comput. Surv.}, 55(12), March 2023.

\bibitem{Legg2007ACO}
Shane Legg and Marcus Hutter.
\newblock A collection of definitions of intelligence.
\newblock In {\em Artificial General Intelligence}, 2007.

\bibitem{Ouyang2024}
Long Ouyang, Jeff Wu, Xu~Jiang, Diogo Almeida, Carroll~L. Wainwright, Pamela
  Mishkin, Chong Zhang, Sandhini Agarwal, Katarina Slama, Alex Ray, John
  Schulman, Jacob Hilton, Fraser Kelton, Luke Miller, Maddie Simens, Amanda
  Askell, Peter Welinder, Paul Christiano, Jan Leike, and Ryan Lowe.
\newblock Training language models to follow instructions with human feedback.
\newblock In {\em Proceedings of the 36th International Conference on Neural
  Information Processing Systems}, NIPS '22, Red Hook, NY, USA, 2024. Curran
  Associates Inc.

\bibitem{Petroni2019}
Fabio Petroni, Tim Rockt{\"a}schel, Patrick Lewis, Anton Bakhtin, Yuxiang Wu,
  AH~Miller, and Sebastian Riedel.
\newblock Language models as knowledge bases?
\newblock 01 2019.

\bibitem{Popper1968}
K.R. Popper.
\newblock Epistemology without a knowing subject.
\newblock In B.~{Van Rootselaar} and J.F. Staal, editors, {\em Logic,
  Methodology and Philosophy of Science III}, volume~52 of {\em Studies in
  Logic and the Foundations of Mathematics}, pages 333--373. Elsevier, 1968.

\bibitem{Russell1918}
Bertrand Russell.
\newblock The philosophy of logical atomism.
\newblock {\em The Monist}, 29(3):345--380, 1918.

\bibitem{Ryle1946}
G.~Ryle.
\newblock Knowing how and knowing that.
\newblock In {\em Collected Papers (Volume 2)}, pages 212--225. New York:
  Barnes and Nobles, 1971.
\newblock First published 1946.

\bibitem{shortliffe2012computer}
Edward Shortliffe.
\newblock {\em Computer-based medical consultations: MYCIN}, volume~2.
\newblock Elsevier, 2012.

\bibitem{Sterkenburg2021}
Tom~F. Sterkenburg and Peter~D. Gr{\"u}nwald.
\newblock The no-free-lunch theorems of supervised learning.
\newblock {\em Synthese}, 199(3):9979--10015, 2021.

\bibitem{vandenBosch2005ScalableCW}
Antal van~den Bosch.
\newblock Scalable classification-based word prediction and confusible
  correction.
\newblock {\em International Journal of Web Engineering and Technology}, 2005.

\bibitem{Vaswani2017}
Ashish Vaswani, Noam Shazeer, Niki Parmar, Jakob Uszkoreit, Llion Jones,
  Aidan~N Gomez, \L~ukasz Kaiser, and Illia Polosukhin.
\newblock Attention is all you need.
\newblock In I.~Guyon, U.~Von Luxburg, S.~Bengio, H.~Wallach, R.~Fergus,
  S.~Vishwanathan, and R.~Garnett, editors, {\em Advances in Neural Information
  Processing Systems}, volume~30. Curran Associates, Inc., 2017.

\bibitem{Wei2024}
Jason Wei, Xuezhi Wang, Dale Schuurmans, Maarten Bosma, Brian Ichter, Fei Xia,
  Ed~H. Chi, Quoc~V. Le, and Denny Zhou.
\newblock Chain-of-thought prompting elicits reasoning in large language
  models.
\newblock In {\em Proceedings of the 36th International Conference on Neural
  Information Processing Systems}, NIPS '22, Red Hook, NY, USA, 2024. Curran
  Associates Inc.

\bibitem{Wittgenstein_PI}
Ludwig Wittgenstein.
\newblock {\em Philosophical Investigations}.
\newblock Blackwell Publishing, 1953.

\bibitem{9010283}
Xiaohua Zhai, Avital Oliver, Alexander Kolesnikov, and Lucas Beyer.
\newblock S4l: Self-supervised semi-supervised learning.
\newblock In {\em 2019 IEEE/CVF International Conference on Computer Vision
  (ICCV)}, pages 1476--1485, 2019.

\bibitem{Zhang2023}
L.~Zhang, X.~Zhai, Z.~Zhao, X.~Wen, and B.~Zhao.
\newblock What if the tv was off? examining counterfactual reasoning abilities
  of multi-modal language models.
\newblock In {\em 2023 IEEE/CVF International Conference on Computer Vision
  Workshops (ICCVW)}, pages 4631--4635, Los Alamitos, CA, USA, oct 2023. IEEE
  Computer Society.

\end{thebibliography}
\end{document}